\documentclass[sigconf]{acmart}
\AtBeginDocument{%
  }

\setcopyright{acmlicensed}
\copyrightyear{2018}
\acmYear{2018}
\acmDOI{XXXXXXX.XXXXXXX}
\acmConference[Conference acronym 'XX]{Make sure to enter the correct
  conference title from your rights confirmation email}{June 03--05,
  2018}{Woodstock, NY}
\acmISBN{978-1-4503-XXXX-X/2018/06}

\usepackage{booktabs}
\usepackage{multirow}
\usepackage{makecell}

\begin{document}

\title{Think over Trajectories: Leveraging Video Generation to Reconstruct GPS Trajectories from Cellular Signaling}


\author{Ruixing Zhang}
\affiliation{%
  \institution{The State Key Laboratory of Complex and Critical Software Environment, Beihang University}
  \country{China}}
\email{yyxzhj@buaa.edu.cn}

\author{Hanzhang Jiang}
\affiliation{%
  \institution{The State Key Laboratory of Complex and Critical Software Environment, Beihang University}
  \country{China}}
\email{21375212@buaa.edu.cn}

\author{Leilei Sun}
\affiliation{%
  \institution{The State Key Laboratory of Complex and Critical Software Environment, Beihang University}
  \institution{H3I, Beihang University}
  \country{China}
  }
\email{leileisun@buaa.edu.cn}

\author{Liangzhe Han}
\affiliation{%
  \institution{The State Key Laboratory of Complex and Critical Software Environment, Beihang University}
  \country{China}}
\email{liangzhehan@buaa.edu.cn}

\author{Jibin Wang}
\affiliation{
  \institution{China Mobile Information Technology Center}
  \country{China}}
\email{wangjibin@chinamobile.com}

\author{Weifeng Lv}
\affiliation{%
  \institution{The State Key Laboratory of Complex and Critical Software Environment, Beihang University}
  \country{China}
}
\email{lwf@buaa.edu.cn}

\renewcommand{\shortauthors}{Trovato et al.}

\begin{abstract}
Mobile devices continuously interact with cellular base stations, generating massive volumes of signaling records that provide broad coverage for understanding human mobility.
However, such records offer only coarse location cues (e.g., serving-cell identifiers) and therefore limit their direct use in applications that require high-precision GPS trajectories.
This paper studies the Sig2GPS problem: reconstructing GPS trajectories from cellular signaling.
Inspired by domain experts often lay the signaling trace on the map and sketch the corresponding GPS route,
unlike conventional solutions that rely on complex multi-stage engineering pipelines or regress coordinates, Sig2GPS is reframed as an image-to-video generation task that directly operates in the map-visual domain: signaling traces are rendered on a map, and a video generation model is trained to draw a continuous GPS path.
To support this paradigm, a paired signaling-to-trajectory video dataset is constructed to fine-tune an open-source video model, and a trajectory-aware reinforcement learning-based optimization method is introduced to improve generation fidelity via rewards.
Experiments on large-scale real-world datasets show substantial improvements over strong engineered and learning-based baselines, while additional results on next GPS prediction indicate scalability and cross-city transferability.
Overall, these results suggest that map-visual video generation provides a practical interface for trajectory data mining by enabling direct generation and refinement of continuous paths under map constraints.
\end{abstract}


\maketitle

\section{Introduction}

With the ubiquity of mobile devices, telecom operators can collect increasingly rich cellular signaling records generated through interactions between cell phones and base stations.
According to internal statistics, a user generates about 200 signaling records per day on average and owns 1.3 mobile devices on average.
These records implicitly capture the person-time-location relationship, enabling a wide range of downstream analytics such as mobility modeling, occupation inference, user profiling, and regional insights\cite{UrbanClip, graphst, autost}. 

\begin{figure}[t]
  \centering
  \includegraphics[width=\linewidth]{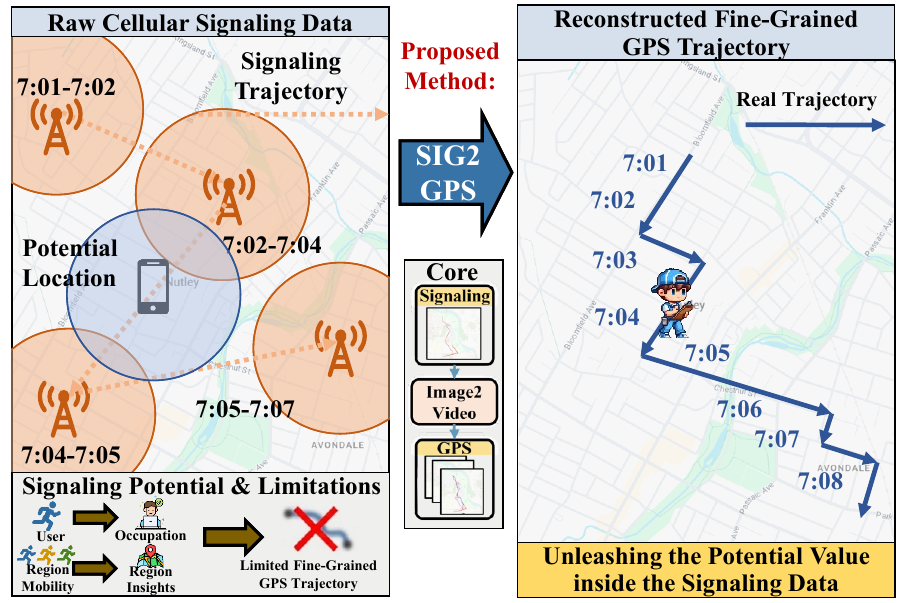}
  \caption{Overview of the Sig2GPS problem: coarse cellular signaling records are mapped to high-precision GPS trajectories via video generation.}
  \label{fig:overview}
\end{figure}

However, a key bottleneck that limits the practical value of cellular signaling data is \textbf{its coarse spatial resolution}: each record typically reflects only the serving base station rather than a user's precise location.
If we could reliably transform wide-coverage signaling records into fine-grained GPS trajectories, it would substantially broaden their applicability and unlock greater value for mobility-centric analytics and services.
The task is illustrated in the Figure~\ref{fig:overview}.

In practice, converting cellular signaling records into GPS trajectories is challenging and typically requires a long pipeline including ping-pong effect mitigation, map matching, and route inference.
Industrial deployments often rely on multi-stage, highly engineered workflows, where each step incurs non-trivial latency.
Moreover, real-world environments are heterogeneous, and extensive case-by-case heuristics are frequently required, resulting in complex codebases that \textbf{depend heavily on expert knowledge and remain difficult to automate at scale}.

Discovered by empirical practice, overlaying signaling trajectories on a map makes it substantially easier to identify the plausible underlying GPS path.
Given a signaling trace and a map, domain experts can often sketch the corresponding route quickly.
However, encoding such tacit, visualization-driven expertise into a robust signal-to-GPS algorithm is non-trivial.

Motivated by recent progress in geometric reasoning with video generation models (e.g., generating valid solutions for maze-like spatial problems), video generation emerges as a promising mechanism for spatially coherent trajectory refinement.
Accordingly, this work casts Sig2GPS as an image-to-video generation problem: a model is conditioned on a map-based visualization of cellular signaling and is tasked with generating a continuous GPS trajectory that is both spatially plausible and consistent with the observations.
This paradigm, termed Think Over Trajectory, aligns with how practitioners reason about signaling data by directly drawing the path on a map.

\begin{figure}[t]
  \centering
  \includegraphics[width=\linewidth]{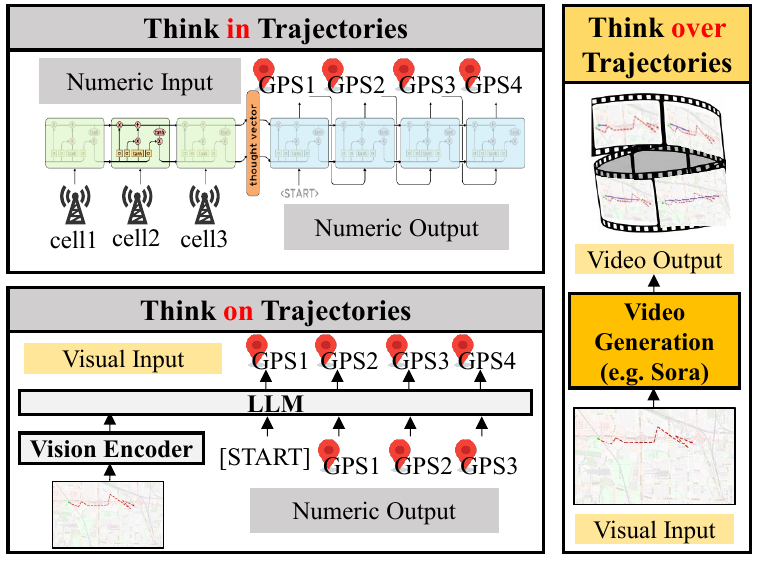}
  \caption{Illustration of Think Over Trajectory. We highlight the difference between previous seq2seq methods, VLM-based methods and our proposed methods.}
  \label{fig2}
\end{figure}

Our framework is different from prior approaches which typically encode GPS trajectories directly.
Alternatively, learning station-specific embeddings can reduce robustness and reusability when base-station deployments change.
Recent VLM-based trajectory mining methods take trajectory visualizations as input, yet they generally output discrete coordinates and do not explicitly model the act of drawing a continuous path on the map, making the coordinate-to-image grounding difficult to learn.
By contrast, Think Over Trajectory explicitly operates in the map-visual domain, providing a more faithful abstraction of human intuition in trajectory reasoning. We illustrates in the difference in the Figure~\ref{fig2}.

To instantiate this paradigm, a paired dataset is constructed that aligns signaling visualizations with GPS-trajectory videos, and an open-source video generation model is fine-tuned to acquire the basic capability of inferring GPS motion from signaling inputs.
Furthermore, since reinforcement learning has been shown to be effective for improving generative quality, a trajectory-aware optimization strategy, Traj-GDPO, is introduced.
The proposed training objective incorporates reward signals that reflect distance error, heading direction, and branching behaviors, enabling further refinement of the generated trajectories.

For evaluation, 10,000 paired signal-GPS trajectories are collected via linkage between two systems.
Experiments demonstrate that the proposed approach substantially outperforms strong industrial baselines and prior learning-based methods.
Moreover, the paradigm is extended to next-GPS prediction, indicating scalability and cross-city transferability. Case studies further show that the generated trajectories adhere to both road-network constraints and signaling observations.

In summary, the main contributions are as follows.

\begin{itemize}
\item \textbf{Paradigm.} A new Think Over Trajectory paradigm is introduced, representing (to the best of current knowledge) the first attempt to leverage video generation for trajectory data mining.
\item \textbf{Method.} We propose Traj-GDPO, a trajectory-aware reinforcement learning fine-tuning method that optimizes video generations with verifiable rewards.
\item \textbf{Results.} Strong empirical performance is achieved on both Sig2GPS Task and next-GPS prediction task, demonstrating favorable scalability, transferability, and generation fidelity.

\end{itemize}

\section{Related Work}

\subsection{Trajectory Data Mining}


Trajectory data provide a primary representation of human mobility and trajectory data mining underpin many real-world applications such as traffic resource optimization, epidemic forecasting, and urban mobility management.

Early approaches modeled human mobility using Markov chainss\cite{MarkovChains}, which capture low-order transitions. With the rise of deep learning, recurrent neural networks\cite{RNN} and Transformer-based\cite{Transformer} architectures became widely adopted for trajectory modeling and prediction. Subsequent studies to mine geographic context via graph-based learning and spatial feature embeddings\cite{unimob, getnext}. More recently, large language model (LLM)-based approaches (e.g., LLM-Mob\cite{jiang}, Agent-Move\cite{agentmove} and NextLocLLM\cite{nextlocllm}) explored leveraging semantic priors and textual knowledge to enhance mobility reasoning.
Overall, most existing methods treat trajectories primarily as numeric sequences or discrete symbols (i.e., reasoning “on” trajectories), and their outputs remain in coordinate or token form. 

With the development of vision-language models (VLMs), trajectory mining has also been studied through visual representations, providing richer contextual cues such as TrajMLLM\cite{trajmllm} and VGLS\cite{vgls}. Nevertheless, outputs are still typically expressed as numeric coordinates rather than continuous paths drawn in the visual domain.

\subsection{Video Generation}


Video generation has progressed along a clear line of probabilistic modeling: VAE-\cite{vae} $\rightarrow$ diffusio\cite{diffusion} $\rightarrow$ Flow Matching\cite{flowmatching}. Compared to VAE and Diffusion models, continuous-time flow-based generators, especially Flow Matching, cast generation as integrating a learned velocity field and often enable fewer-step sampling, making them a natural substrate for structured refinement.

Recent large-scale video models further suggest that video generators can exhibit reasoning behaviors over images. In particular, Google's veo3\cite{veo3} has demonstrated it can solve a broad variety of tasks it wasn't meant to train for such as segmenting objects, detecting edges, editing images, and much more. These abilities motivate our choice to formulate Sig2GPS as map-visual video generation, leveraging spatiotemporal reasoning to produce topology-consistent continuous paths instead of unconstrained coordinate regression.

\subsection{Reinforcement Learning from Verifiable Reward}

Reinforcement learning from verifiable reward (RLVR) has emerged as a practical mechanism for improving large models by exploiting feedback that is programmatically checkable. Following early successes such as GPT-o1\cite{o1}, subsequent work has shown that interaction-based self-improvement with outcome supervision (e.g., correctness on reasoning tasks) can yield substantial gains, inspiring reasoning-oriented systems such as Deepseek R1\cite{deepseekr1}. Most RLVR pipelines build on policy optimization methods such as PPO\cite{ppo}, while recent approaches adopt GRPO\cite{grpo} as a lightweight alternative that replaces a learned value function with group-relative advantages.

RLVR has also been extended to image and video generation, where rewards can be defined by domain constraints or automatic evaluators. Representative directions include Flow-GRPO\cite{flowgrpo}, Dance-GRPO\cite{dancegrpo}, and Dense-GRPO\cite{densegrpo}, which leverage the ODE-to-SDE connection in flow-based models to enable multiple rollouts per condition and stable group-relative updates. These advances suggest that reinforcement learning can complement supervised training by directly optimizing for task-aligned criteria that are difficult to express via likelihood objectives.

\section{Preliminaries}

\subsection{Problem Defination}

\textbf{Cellular Signaling Data}: In this study, the original cellular signaling data is defined as a temporal sequence of cellular station connections, represented as $ T = \{T_{1}, T_{2}, \ldots, T_{|T|}\} $, where each element $ T_{j} = \{x_{j}, y_{j}, t^0_{j}, t^1_{j}\} $ signifies the mobile phone's connection to a cellular station located at $(x_{j}, y_{j})$ from $ t^0_{j} $ to $ t^1_{j} $. 

\textbf{GPS Trajectory Data}: The ground-truth GPS trajectory is defined as an evenly-sampled sequence $ G = \{G_{1}, G_{2}, \cdots, G_{|G|} \}$, where each point $ G_{k} = \{\hat{x}_{k},\hat{y}_{k},\hat{t}_{k}\}$ denotes the GPS coordinate at timestamp $\hat{t}_{k}$. The fixed sampling interval is assumed as $\Delta t$, i.e., $\hat{t}_{k+1}-\hat{t}_{k}=\Delta t$.

\textbf{Sig2GPS Task}: Given a cellular signaling sequence $T$ (and the corresponding map context), Sig2GPS aims to reconstruct the corresponding fine-grained GPS trajectory $G$.
Formally, we learn a mapping $\mathcal{F}_\theta$ such that $\hat{G} = \mathcal{F}_\theta(T)$, and $\hat{G}$ is as close as possible to the ground truth $G$ while remaining spatially plausible under map constraints. We align the start time of the GPS trajectory with that of the signaling sequence and resample GPS points at a fixed interval of $\Delta t=15\,\mathrm{s}$.

\begin{figure*}[t]
  \centering
  \includegraphics[width=0.95\textwidth]{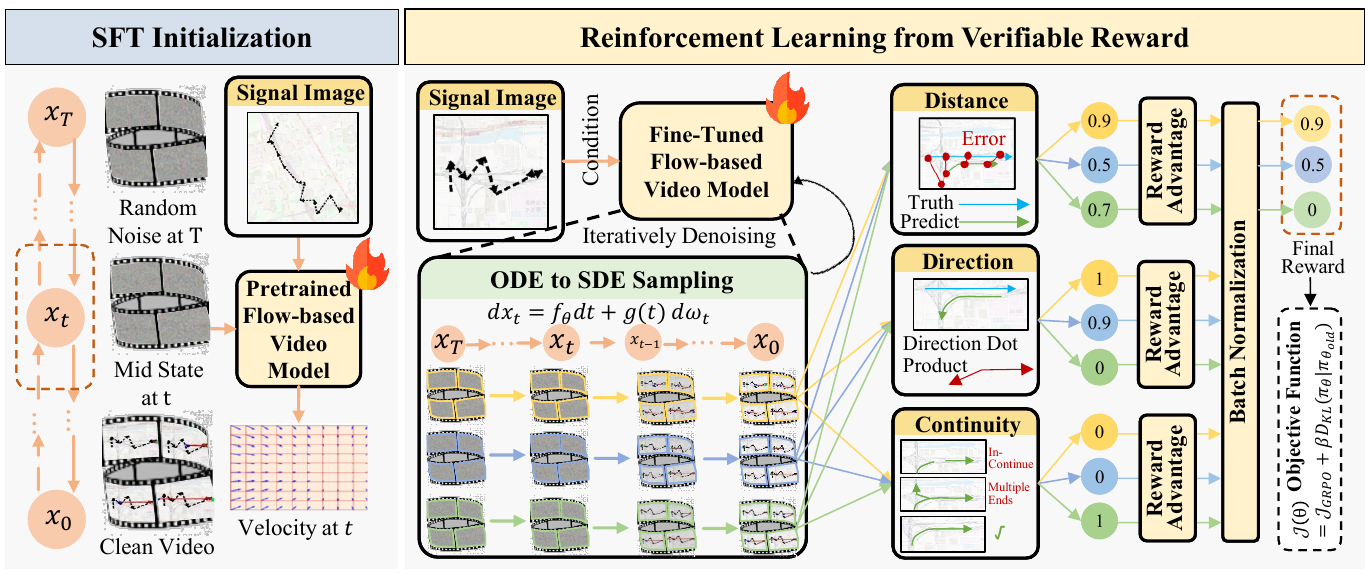}
  \caption{Overall architecture of our Think Over Trajectory framework for Sig2GPS. We first fine-tune the video generator on paired signaling-trajectory videos, and then apply trajectory-aware reinforcement learning from verifiable reward (RLVR) to further improve fidelity under map and temporal constraints. }
  \label{fig:model}
\end{figure*}

\subsection{Reinforcement Learning on Flow-based Model}

Recent video generators are increasingly implemented as flow-based generative models, where a sample is produced by integrating a learned vector field over a continuous time variable.
Let $x_t \in \mathbb{R}^d$ denote a latent variable at time $t \in [0,1]$. A conditional flow model defines an ordinary differential equation (ODE)
\begin{equation}
\frac{\mathrm{d}x_t}{\mathrm{d}t} = f_\theta(x_t, t \mid c),
\end{equation}
where $c$ denotes conditioning information (e.g., the signaling visualization) and $f_\theta$ is parameterized by a neural network.
Integrating this probability flow from an initial noise distribution $p_0$ yields a sample $x_1$ that can be decoded into a trajectory video.

However, the deterministic ODE formulation can be inconvenient for RL-style optimization, since group-based estimators typically require multiple stochastic rollouts to compute relative advantages, whereas integrating the same ODE under fixed conditions yields identical trajectories. To connect the ODE formulation with diffusion-style stochasticity, the corresponding stochastic differential equation (SDE) can be considered:
\begin{equation}
\mathrm{d}x_t = f_\theta(x_t, t \mid c)\,\mathrm{d}t + g(t)\,\mathrm{d}w_t,
\end{equation}
where $w_t$ is a standard Wiener process and $g(t)$ controls the noise schedule.
Under mild regularity conditions, the marginal density evolution induced by the SDE can be matched by a deterministic probability flow ODE with drift
$f_\theta(x_t,t\mid c) - \tfrac{1}{2} g(t)^2 \nabla_{x} \log p_t(x_t\mid c)$.
This ODE-SDE equivalence motivates interpreting conditional generation as controlling a continuous-time dynamical system whose terminal state determines the generated video.

When a single generation is donated as a rollout $\tau$,  a common KL-regularized objective can be written as
\begin{equation}
\max_{\theta} \; \mathbb{E}_{\tau \sim p_{\theta}(\cdot\mid c)}\big[R(\tau)\big] - \beta\,\mathrm{KL}\big(p_{\theta}(\cdot\mid c)\,\|\,p_{\theta_0}(\cdot\mid c)\big),
\end{equation}
where $\theta_0$ denotes the supervised-initialized model.

To obtain a stable update without requiring an additional model, a group-relative policy optimization (GRPO) style estimator can be naturally adopted.
Specifically, for each conditioning input $c$, a group of $K$ rollouts $\{\tau_i\}_{i=1}^{K}$ is sampled from a frozen reference policy $p_{\theta_{\mathrm{old}}}(\cdot\mid c)$.
Group-relative advantages are computed by normalizing rewards within the group,
\begin{equation}
A_i = \frac{R(\tau_i) - \frac{1}{K}\sum_{j=1}^{K} R(\tau_j)}{\mathrm{Std}\big(\{R(\tau_j)\}_{j=1}^{K}\big) + \epsilon},
\end{equation}
Then, following a clipped policy-gradient update, the GRPO objective is
\begin{equation}
\begin{aligned}
\mathcal{L}_{\mathrm{GRPO}}(\theta)=&
\mathbb{E}\left[\frac{1}{K}\sum_{i=1}^{K}
\min\Big(r_i(\theta)A_i,\; \mathrm{clip}(r_i(\theta),1-\delta,1+\delta)A_i\Big)
\right] \\
-& \beta\,\mathrm{KL}\big(p_{\theta}(\cdot\mid c)\,\|\,p_{\theta_{\mathrm{old}}}(\cdot\mid c)\big),
\end{aligned}
\end{equation}
where $r_i(\theta)=\exp\big(\log p_{\theta}(\tau_i\mid c)-\log p_{\theta_{\mathrm{old}}}(\tau_i\mid c)\big)$.
In the flow-based setting, $\log p_{\theta}(\tau\mid c)$ is the model's path probability.

\section{Methodology}
\label{sec:method}

Our framework is illustrated at Figure~\ref{fig:model}, which follows a two-stage training recipe: (i) supervised fine-tune (SFT) a flow-based video generator on paired signaling-trajectory videos. (ii) Trajectory-aware reinforcement learning from verifiable rewards to further align generations with map topology and temporal consistency.

\subsection{Pair Data Collection}
\label{sec:pair}

\paragraph{Data sources.}
We leverage two complementary data streams collected by a telecom operator and its mobility ecosystem.
The first stream is cellular signaling, i.e., time-stamped associations between a device and serving cell towers.
The second stream is high-frequency taxi GPS trajectories recorded by a fleet management platform.
Taxi GPS provides a reliable approximation of on-road movement and serves as the supervision signal.

\paragraph{Cross-system linkage}
Due to personal privacy, direct links are not available across the two systems.
Instead, we construct paired samples by matching signaling sequences with taxi trajectories using spatiotemporal consistency.
Concretely, for each candidate taxi trajectory $G$, we directly measure how it fits the observed signaling trace $T$ by evaluating a spatiotemporal consistency.

We accept a pair $(T,G)$ when $G$ satisfies three practical but effective criteria: \textbf{(i) mobility}, the taxi trajectory exhibits sustained movement (excluding long parking intervals) during the matched window; \textbf{(ii) time coverage}, the overlapped duration is sufficiently long (e.g., exceeding 6 hours within a day) to avoid incidental matches; and \textbf{(iii) distance consistency}, GPS points remain within a bounded distance to the corresponding serving-tower locations over time.
If multiple candidates satisfy the criteria, we keep only the one with the smallest mismatch and discard the rest to avoid ambiguous supervision.
After these steps, we obtain approximately 20{,}000 high-confidence signaling--taxi pairs.

\paragraph{Map-visual rendering for video supervision.}
To cast Sig2GPS into image-to-video generation, each pair is converted into a training example $(c, y)$:
(1) conditioning input $c$ is a map tile based on OpenStreetMap\footnote{\url{https://www.openstreetmap.org/}} centered at the trace region with a rendered signaling polyline.
(2) target video $y$ is a short sequence in which the ground-truth GPS path is progressively drawn over the same map.

\subsection{SFT Initialization}
\label{sec:sft}

Through the design above, given the conditioning image $c$ (a map tile overlaid with the signaling trace), the model generates a $F$-frame video $\hat{y}$ that draws the GPS path on the map over time.
This map-visual drawing  intentionally mirrors how cellular-signaling engineers reason in practice: rather than manipulating raw coordinate sequences, they inspect signaling footprints on a map and sketch a plausible on-road route constrained by topology.
Compared with purely numeric trajectory modeling, this representation exposes road geometry explicitly. Compared with using a trajectory image merely as an input while predicting discrete coordinates as output, it keeps both conditioning and prediction in a unified visual space, avoiding an extra image-to-number projection that is hard to learn and hard to verify.

Because the open-source video generation model is not inherently trained on Sig2GPS problem, we first fine-tune an open-source video generation model that is implemented based on a flow-based architecture.
Let $y$ denote the rendered ground-truth trajectory video.
We optimize a standard conditional flow-matching objective, which trains the network to predict the velocity field along the probability flow.
Denoting the model by $f_\theta(\cdot)$, the SFT stage minimizes
\begin{equation}
\mathcal{L}_{\mathrm{SFT}}(\theta)=\mathbb{E}_{(c,y)}\,\mathbb{E}_{t\sim\mathcal{U}(0,1)}\big[\ell\big(f_\theta(x_t,t\mid c),\; v_t\big)\big],
\end{equation}
where $x_t$ is the interpolated latent/state at time $t$, $v_t$ is the corresponding target velocity, and $\ell$ is an $\ell_2$ regression loss.

\subsection{Trajectory-aware reinforcement learning from verifiable rewards}

While SFT provides a strong initialization, its training signal is largely driven by image-level reconstruction objectives.
As a result, the model can achieve a low pixel-level loss while still making fine-grained but critical mistakes, e.g., drawing a path with the correct shape yet taking a wrong turn at an intersection, reversing local direction, or violating road topology.
We therefore introduce reinforcement learning from verifiable rewards to explicitly score and optimize these fine-grained trajectory criteria, enabling optimization beyond coarse visual similarity.

\subsubsection{Rewards}

To improve the fine-grained quality of trajectory videos, we design three verifiable rewards by combining (i) the evaluation criteria we ultimately care about and (ii) the typical failure modes we observe from the SFT-initialized generator.

Let $y$ be the ground-truth trajectory video and $\hat{y}$ be the generated one, both containing $F$ frames.
From each frame $m$, we extract the blue dot $p_m\in\mathbb{R}^2$ in the polyline as it represents the end point. Correspondingly, we denote the point in $\hat y$ as $ \hat{p}_m$.

\paragraph{(1) Distance reward.}
This reward directly measures trajectory error at key stages (first frame, last frame, and three intermediate frames).
For each anchor frame $m\in\{1,\lfloor F/4\rfloor,\lfloor F/2\rfloor,\lfloor 3F/4\rfloor,F\}$, we compute the geodesic distance:
\begin{equation}
\mathrm{dis}(m)=d_{\mathrm{geo}}\bigl(\hat{p}_m,\,p_m\bigr).
\end{equation}
We then map it to a bounded reward
\begin{equation}
R_{\mathrm{dist}}=\frac{1}{|\mathcal{M}|}\sum_{m\in\mathcal{M}}\left(1-\frac{\min\{\mathrm{dis}(m),5000\}}{5000}\right).
\end{equation}
Intuitively, this encourages the generator to be close to the target throughout the drawing process.

\paragraph{(2) Direction reward.}
Pixel-level losses can be insensitive to directional mistakes (e.g., a path that looks plausible but is traversed in the wrong direction).
We therefore reward directional alignment using the normalized displacement vectors between the start and end anchors:
\begin{equation}
\mathbf{v}=p_F-p_1,\quad \hat{\mathbf{v}}=\hat{p}_F-\hat{p}_1,
\end{equation}
\begin{equation}
R_{\mathrm{dir}}=\frac{\left\langle \frac{\hat{\mathbf{v}}}{\lVert \hat{\mathbf{v}}\rVert_2+\epsilon},\; \frac{\mathbf{v}}{\lVert \mathbf{v}\rVert_2+\epsilon}\right\rangle + 1}{2}.
\end{equation}
This yields $R_{\mathrm{dir}}\in[0,1]$, where $1$ indicates perfectly consistent direction.

\paragraph{(3) Continuity reward.}
We penalize broken or fragmented generations.
After extracting the final-frame trajectory from $\hat{y}$, we compute the number of connected components $C$ and the number of endpoints $E$ (pixels with degree $1$ under 8-neighborhood connectivity).
The reward is defined as a hard constraint
\begin{equation}
R_{\mathrm{cont}}=\mathbb{I}[C=1\ \wedge\ E=2],
\end{equation}
which equals $0$ if multiple start/end points or multiple disconnected regions are detected, and equals $1$ otherwise.

Finally, we set the reward vector as $\mathbf{r}(\tau)=[R_{\mathrm{dist}},R_{\mathrm{dir}},R_{\mathrm{cont}}]$.

\subsubsection{Traj-GDPO}

Our reward is naturally multi-component, but different criteria (e.g., map compliance vs.
 signaling consistency) have different numerical ranges, sparsity patterns, and noise levels.
A naive scalarization (fixed weighted sum) can cause training instability: a large-range reward dominates the advantage estimate, while a small-range but important reward becomes ineffective.
To address this, inspired by GDPO\cite{gdpo}, we propose Trajectory-aware Group Decoupled Policy Optimization (Traj-GDPO), a group-relative optimization procedure that normalizes and aggregates heterogeneous rewards.

\paragraph{Group sampling.}
For each conditioning input $c_i$ in a mini-batch $\mathcal{D}_{\mathrm{Batch}}=\{c_i\mid i=1,\ldots,B\}$, we sample a group of $K$ rollouts from the current policy, obtaining trajectories $\{\tau^{(i,k)}\}_{k=1}^{K}$.
Each rollout is evaluated by $M$ reward components (in our case $M=3$), denoted as $r_m^{(i,k)}\in\mathbb{R}$ for reward index $m\in\{1,\ldots,M\}$.

\paragraph{Decoupled normalization}
We perform normalization separately for each reward component to decouple heterogeneous reward scales.
For each $i$ and reward component $m$, we compute the group-relative advantage
\begin{equation}
A_m^{(i,k)}=
\frac{r_m^{(i,k)}-\mathrm{mean}_{k'=1}^{K}\,r_m^{(i,k')}}{\mathrm{std}_{k'=1}^{K}\,r_m^{(i,k')}+\epsilon}.
\end{equation}
We then aggregate objectives by summation,
\begin{equation}
A_{\mathrm{sum}}^{(i,k)}=\sum_{m=1}^{M} A_m^{(i,k)}.
\end{equation}

\paragraph{Batch normalization for numerical stability.}
To keep a stable numerical range across updates, GDPO further normalizes $A_{\mathrm{sum}}^{(i,k)}$ over the full mini-batch:
\begin{equation}
\hat{A}_{\mathrm{sum}}^{(i,k)}=
\frac{A_{\mathrm{sum}}^{(i,k)}-\mathrm{mean}_{i'\in\mathcal{D}_{\mathrm{Batch}},\,k'\in\{1,\ldots,K\}}\,A_{\mathrm{sum}}^{(i',k')}}{\mathrm{std}_{i'\in\mathcal{D}_{\mathrm{Batch}},\,k'\in\{1,\ldots,K\}}\,A_{\mathrm{sum}}^{(i',k')}+\epsilon}.
\end{equation}

\paragraph{Optimization objective.}
We adopt a clipped policy optimization objective with KL regularization, using $\hat{A}_{\mathrm{sum}}^{(i,k)}$ as the final advantage.
In our setting, we observe a clear degeneration phenomenon during RL: if the KL penalty is removed as Dr.GRPO\cite{drgrpo}, or if the KL reference is chosen as the previous policy $p_{\theta_{\mathrm{old}}}$, the generated videos tend to gradually collapse (e.g., producing overly similar and less informative trajectories) as training proceeds.
We therefore keep the importance ratio against $p_{\theta_{\mathrm{old}}}$ for a stable clipped update, while anchoring the KL term to a fixed initialization policy $p_{\theta_{\mathrm{initial}}}$ (the SFT checkpoint) to prevent drifting away from the supervised solution manifold.
Let the likelihood ratio be
\begin{equation}
\rho^{(i,k)}(\theta)=\exp\big(\log p_{\theta}(\tau^{(i,k)}\mid c^{(i)})-\log p_{\theta_{\mathrm{old}}}(\tau^{(i,k)}\mid c^{(i)})\big).
\end{equation}
Then we maximize
\begin{equation}
\begin{aligned}
&\mathcal{L}_{\mathrm{Traj\text{-}GDPO}}(\theta)=
\mathbb{E}\Bigg[\frac{1}{BK}\sum_{i=1}^{B}\sum_{k=1}^{K}
\min\Big(\rho^{(i,k)}(\theta)\hat{A}_{\mathrm{sum}}^{(i,k)},\; 
\\
&\mathrm{clip}(\rho^{(i,k)}(\theta),1-\delta,1+\delta)\hat{A}_{\mathrm{sum}}^{(i,k)}\Big)\Bigg]
-\beta\,\mathrm{KL}\big(p_{\theta}(\cdot\mid c)\,\|\,p_{\theta_{\mathrm{initial}}}(\cdot\mid c)\big).
\end{aligned}
\end{equation}

\subsection{Effect of Roll-Out}

\begin{figure}[t]
  \centering
  \includegraphics[width=\linewidth]{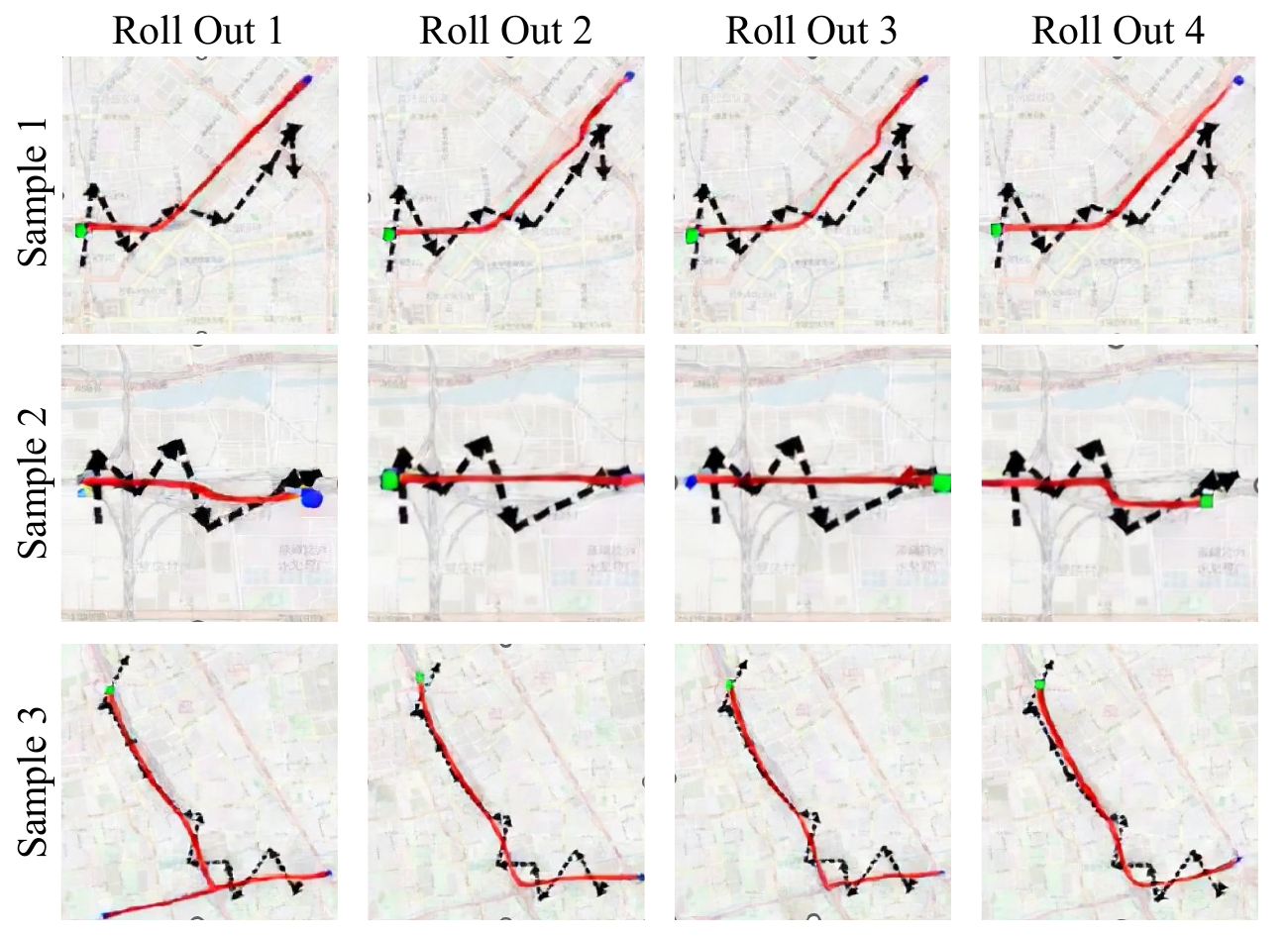}
  \caption{Three Examples of Our Roll Out.}
  \label{fig:rollout}
\end{figure}

Figure~\ref{fig:rollout} visualizes diverse rollouts sampled during RL training via the proposed ODE-to-SDE sampling scheme, where green indicates the start point and blue indicates the end point.
The rollouts illustrate that the generator explores multiple plausible routing hypotheses under the same conditioning input.

For example, in Sample 1, different rollouts correspond to alternative on-road choices such as traversing the inner lane directly, taking an outer-lane detour that merges into a service road, or following a slightly different lane-level path.
Similarly, in Sample 2, some rollouts initiate from the left branch (Rollouts 1--2), while others initiate from the right branch (Rollouts 3--4), highlighting the benefit of generative modeling in producing multiple reasonable candidates.
In Sample 3, Rollout~1 produces two plausible trajectory segments within a single run, which motivates the introduction of the continuity reward.
This diversity is essential for reinforcement learning, as it enables credit assignment by promoting high-reward trajectories among sampled alternatives.

\section{Experiments}

\subsection{Experimental Settings}

\subsubsection{Model Settings}

Wan2.2-TI2V-5B\cite{wan} is used as the base video generator, as it is one of the strongest open-source image-to-video models available.
In the main setting, $F=21$ frames are generated per sample.
The generator is fine-tuned with LoRA in both SFT and RL stages, using rank $r=28$.
Experiments are conducted on four NVIDIA RTX 4090 GPUs: three GPUs for training and one GPU for KL computation against $p_{\theta_{\mathrm{initial}}}$.
Mean absolute error (MAE) and root mean squared error (RMSE) of map-distance errors (in meters) are reported over all predicted points. L100 denotes the proportion of points with error below $100\,\mathrm{m}$, and G1000 denotes the proportion of points with error above $1000\,\mathrm{m}$.

\subsubsection{Dataset}

The Sig2GPS training is performed on 10{,}000 paired signaling-GPS samples collected via the proposed matching pipeline in Beijing.
For efficiency evaluation, we keep 1{,}000 trajectories in the test set, and split the remaining data into training and validation sets with a ratio of $7{:}2$.
We further partition the dataset into Small, Medium, and Large subsets based on the origin-to-destination distance.

To further evaluate generalization beyond Sig2GPS, Next GPS Prediction is additionally considered as a downstream task.
Given the past $10$ GPS points sampled at a fixed interval ($\Delta T=15\,\mathrm{s}$), the model predicts the next $10$ future steps.
Two widely adopted public benchmarks are used for this task: Chengdu and Xi'an.

\begin{table*}[t]
\centering
\caption{Performance comparison. Lower is better for MAE/RMSE/G1000; higher is better for L100.}
\label{tab:size_comparison}
\setlength{\tabcolsep}{5pt}
\renewcommand{\arraystretch}{1.15}
\begin{tabular}{lcccccccccccc}
\toprule
& \multicolumn{4}{c}{\textbf{Small}} & \multicolumn{4}{c}{\textbf{Medium}} & \multicolumn{4}{c}{\textbf{Large}} \\
\cmidrule(lr){2-5}\cmidrule(lr){6-9}\cmidrule(lr){10-13}
\textbf{Model}
& MAE$\downarrow$ & RMSE$\downarrow$ & L100$\uparrow$ & G1000$\downarrow$
& MAE$\downarrow$ & RMSE$\downarrow$ & L100$\uparrow$ & G1000$\downarrow$
& MAE$\downarrow$ & RMSE$\downarrow$ & L100$\uparrow$ & G1000$\downarrow$ \\
\midrule
GRU
& 402.52 & 475.11 & 5.18\% & 2.81\%
& 489.37 & 626.72 & 5.24\% & 7.35\%
& 647.24 & 829.69 & 2.37\% & 15.06\% \\
GPT
& 358.14 & 428.77 & 7.04\% & 2.07\%
& 473.56 & 636.63 & 6.72\% & 6.63\%
& 655.60 & 842.04 & 3.89\% & 16.59\% \\
MLP
& 513.62 & 635.87 & 4.67\% & 10.57\%
& 648.33 & 846.51 & 4.28\% & 18.13\%
& 827.79 & 1016.15 & 1.26\% & 28.68\% \\
TCN
& 613.97 & 737.26 & 2.63\% & 13.53\%
& 722.73 & 873.25 & 1.60\% & 20.30\%
& 951.42 & 1138.49 & 0.90\% & 36.71\% \\
TrajFormer
& 375.12 & 431.60 & 5.12\% & 1.92\%
& 402.13 & 485.89 & 6.20\% & 4.31\%
& 489.70 & 579.33 & 3.62\% & 15.72\% \\
SigFormer
& 329.49 & 423.25 & \underline{12.96}\% & 2.43\%
& 435.86 & 574.51 & 7.29\% & \underline{6.51}\%
& 585.94 & \underline{744.21} & 3.20\% & \underline{12.31}\% \\
$\text{Rule}_{sig}$
& \underline{306.12} & \underline{383.94} & 11.56\% & \underline{2.07}\%
& \underline{403.05} & \underline{530.30} & \underline{10.72}\% & 6.87\%
& \underline{516.80} & 783.21 & \underline{5.60}\% & 18.14\% \\
\midrule
Ours
& \textbf{214.96} & \textbf{307.85} & \textbf{36.71\%} & \textbf{1.02\%}
& \textbf{268.26} & \textbf{367.31} & \textbf{25.57\%} & \textbf{1.62\%}
& \textbf{441.10} & \textbf{701.93} & \textbf{15.24\%} & \textbf{6.99\%} \\
\bottomrule
\end{tabular}
\end{table*}

\subsubsection{Baselines}

For Sig2GPS, comparisons are made against classic deep learning baselines, including GRU, MLP, TCN\cite{TCN}, Transformer-Based Baselines including GPT\cite{GPT3}, TrajFormer\cite{TrajFormer}, SigFormer\cite{SigFormer}, as well as the strongest industrial baseline in our system named $\text{Rule}_{sig}$.
$\text{Rule}_{sig}$ is a composite engineering pipeline that combines Kalman filtering, map matching, and Markov-chain-based route inference.
For next-GPS-location prediction, benchmarking is conducted against established methods such as DeepMove\cite{deepmove}, GetNext\cite{getnext}, UniMob\cite{unimob}, AgentMove\cite{agentmove}, STAN\cite{stan}, and TrajMLLM\cite{trajmllm}.

\subsection{Comparion Results on Sig2GPS}

Table~\ref{tab:size_comparison} summarizes the performance of all methods under three trajectory scope settings. Across all scopes and evaluation metrics, our proposed model consistently achieves the best overall performance. In particular, our model yields the lowest MAE and RMSE while simultaneously achieving the highest L100 and lowest G1000 across all three scopes, demonstrating a clear and robust advantage over recent methods. These results indicate that our model not only improves average reconstruction accuracy, but also significantly reduces large-error cases, as reflected by the improvements on L100 and G1000.

Although the industry-level baseline $\text{Rule}_{sig}$ achieves competitive performance and often ranks second across multiple metrics, it relies on a complex multi-stage pipeline in practice. Specifically, the industrial solution requires KDTree and Kaiman Filter-based ping-pong noise removal, road network matching, and route navigation, with substantial intermediate data transformations between stages. As a result, processing a single trajectory typically takes more than two minutes. In contrast, our proposed Sig2GPS adopts a fundamentally one-step paradigm. By directly generating trajectories that are naturally aligned with the road network, our method eliminates the need for explicit post-processing and map-matching stages. Consequently, Sig2GPS only requires a single image-generation step to enhance signaling quality, and processes a single trajectory within 30 seconds in practice. \textbf{This significant reduction in inference latency highlights the efficiency advantage of our one-step design and suggests strong potential for reshaping future industrial Sig2GPS systems.}

\subsection{Test-Time Scaling}
\begin{table}[h]
\centering
\setlength{\tabcolsep}{4pt}
\caption{Test-time scaling behavior with different numbers of input frames.
Lower is better for both MAE and RMSE.}
\label{tab:test_time_scaling}
\setlength{\tabcolsep}{6pt}
\renewcommand{\arraystretch}{1.15}

\begin{tabular}{c c@{\hspace{3pt}}c c@{\hspace{3pt}}c c@{\hspace{3pt}}c}
\toprule
\multirow{2}{*}{\makecell[c]{\textbf{Frame}\\[-1pt]\textbf{Count}}}
& \multicolumn{2}{c}{\textbf{Small}}
& \multicolumn{2}{c}{\textbf{Medium}}
& \multicolumn{2}{c}{\textbf{Large}} \\
\cmidrule(lr){2-3}
\cmidrule(lr){4-5}
\cmidrule(lr){6-7}
& MAE$\downarrow$ & RMSE$\downarrow$
& MAE$\downarrow$ & RMSE$\downarrow$
& MAE$\downarrow$ & RMSE$\downarrow$ \\
\midrule
13
& 239.64 & 323.69
& 323.83 & \underline{428.35}
& 600.14 & 919.35 \\
17
& \underline{230.20} & \underline{318.49}
& \underline{318.49} & 440.20
& \underline{546.76} & \underline{801.46} \\
21
& \textbf{214.96} & \textbf{268.26}
& \textbf{307.85} & \textbf{367.31}
& \textbf{441.10} & \textbf{761.93} \\
\bottomrule
\end{tabular}
\end{table}
Table~\ref{tab:test_time_scaling} reports the performance of our video-based reasoning model under different numbers of input frames at inference time on Sig2GPS task. 
By increasing the frame count from 13 to 21, we observe a \textbf{consistent reduction in both MAE and RMSE} across three scopes. 
This monotonic improvement indicates that the model can effectively leverage additional temporal context at test time, rather than relying on a fixed-length representation. 
Moreover, the \textbf{performance gains are more pronounced for larger trajectory scope}, suggesting that longer input sequences are particularly beneficial for more complex motion patterns. 
Overall, these results demonstrate a clear test-time scaling behavior without requiring retraining or architectural changes.

\subsection{Comparison Results on Next GPS Prediction and Cross City Transfer}
\begin{table}[h]
\centering
\caption{Performance Comparison on Next GPS Prediction.}
\label{tab:gps}
\begin{tabular}{l cc cc}
\toprule
\multirow{2}{*}{\textbf{Method}} & \multicolumn{2}{c}{\textbf{Chengdu}} & \multicolumn{2}{c}{\textbf{Xi'an}} \\
\cmidrule(lr){2-3}\cmidrule(lr){4-5}
& MAE$\downarrow$ & RMSE$\downarrow$ & MAE$\downarrow$ & RMSE$\downarrow$ \\
\midrule
DeepMove           & 285.36 & 421.41 & 486.52 & 562.46 \\
GetNext      & 239.43 & 316.35 & 577.96 & 661.53 \\
UniMob       & 239.43 & 354.40 & 416.56 & \underline{485.47} \\
AgentMove         & \underline{221.15} & 323.90 & 428.81 & 498.34 \\
STAN          & 257.19 & 377.90 & 455.15 & 523.71 \\
TrajMLLM      & 236.13 & 347.28 & \underline{389.38} & 527.90 \\
Ours (Chengdu) & \textbf{217.70} & \textbf{272.89} & 427.75 & 504.94 \\
Ours (Xi'an)   & 234.85 & \underline{285.35} & \textbf{367.16} & \textbf{446.83} \\
\bottomrule
\end{tabular}
\end{table}
To further evaluate the scalability of our approach beyond trajectory reconstruction, in Table~\ref{tab:gps}, we conduct experiments on the Next GPS Prediction task. Here, Ours (Chengdu) denotes the model trained on the Chengdu dataset, and Ours (Xi'an) denotes the model trained on the Xi'an dataset. As shown by the results, our proposed Think Over Trajectory method consistently outperforms both think-in-trajectory approaches, such as DeepMove, AgentMove, and UniMob, and think-on-trajectory methods like TrajMLLM. \textbf{This performance gap indicates that directly reasoning and generating trajectories in the trajectory-image space provides stronger modeling capacity than operating on discrete trajectory points or static trajectory representations.} 

Moreover, our method demonstrates strong cross-city transferability. Since the model is trained on trajectories aligned with road-network maps, it naturally captures structural priors that are shared across different cities. As a result, models trained on one city can generalize effectively to another. Notably, the model trained on Xi’an even achieves lower RMSE on the Chengdu dataset than all baseline methods, highlighting the robustness and practical potential of our approach. These findings suggest that directly drawing trajectories on road-network-aware images offers a promising direction for trajectory data mining.

\subsection{Ablation Study}

\begin{figure}[h]
  \centering
  \includegraphics[width=0.9\linewidth]{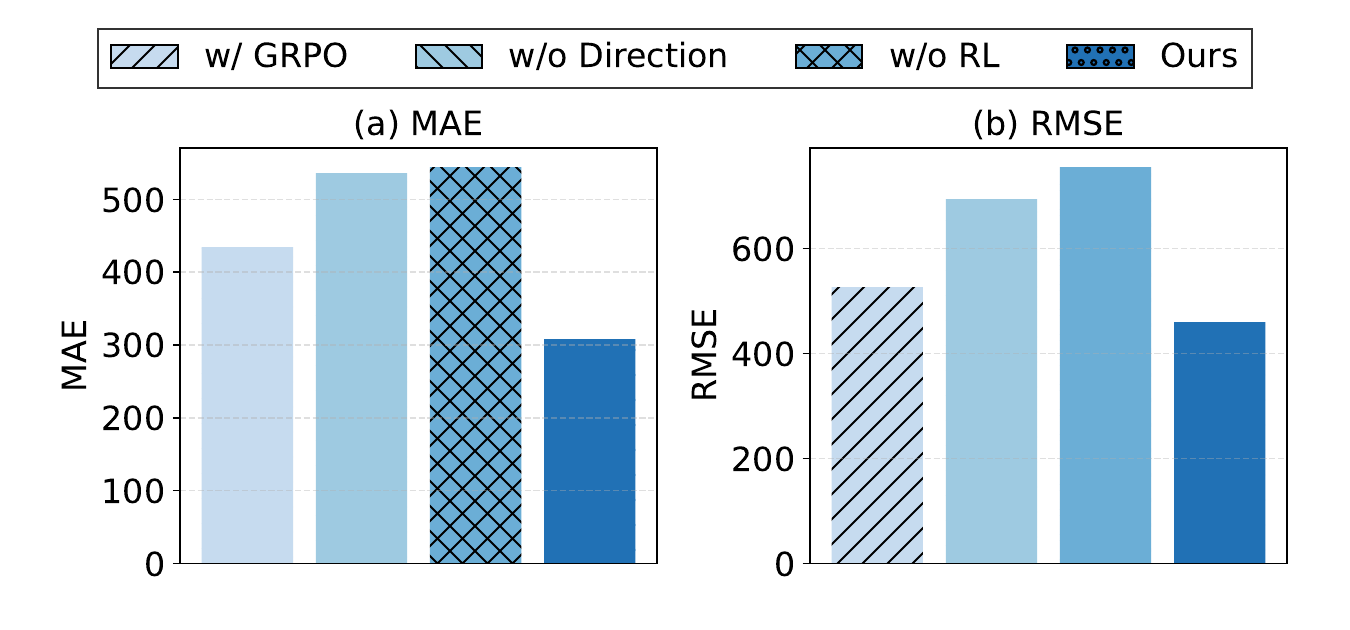}
  \caption{Ablation results on Sig2GPS.}
  \label{fig:ablation}
\end{figure}

Figure~\ref{fig:ablation} reports an ablation study to quantify the impact of each design choice.
Here, w/ GRPO uses GRPO instead of GDPO, and w/o Direction and w/o RL remove $R_{\mathrm{dir}}$ and the whole RL stage respectively.
First, we notice removing Traj-GDPO and training with SFT only consistently degrades performance, indicating that likelihood-based fine-tuning alone is insufficient to correct fine-grained errors.
Second, disabling any individual reward leads to measurable regression: the direction reward reduces directionally incorrect yet visually plausible generations, and empirically accounts for the majority of the performance gains among the three rewards.
Third, replacing GDPO with GRPO (w/ GRPO) still yields slight performance drop, indicating that GRPO is a viable alternative for this task, GDPO is better in our setting.

\begin{figure}[h]
  \centering
  \includegraphics[width=0.9\linewidth]{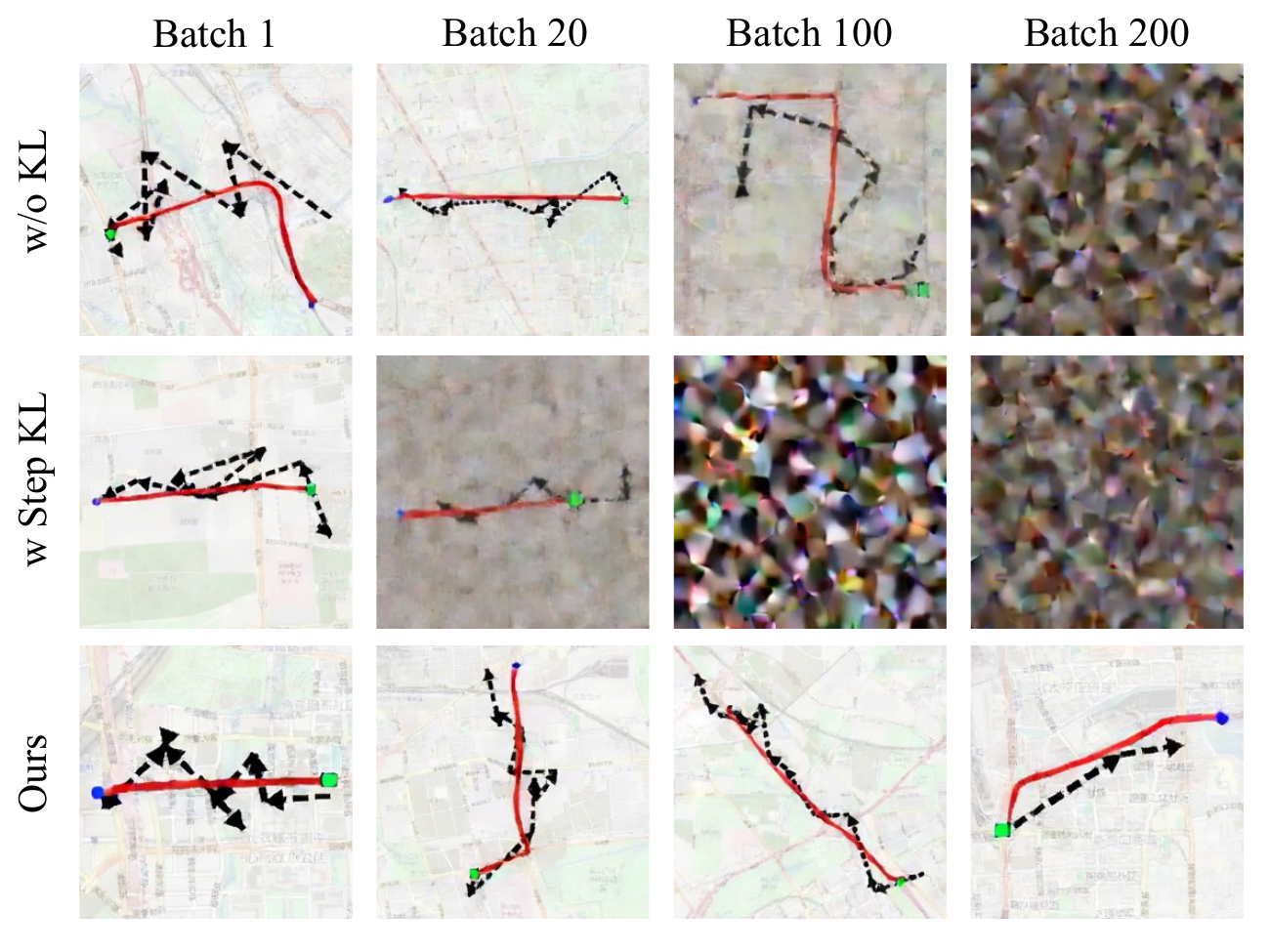}
  \caption{Effect of KL regularization strategies.}
  \label{fig:kl_ablation}
\end{figure}

Figure~\ref{fig:kl_ablation} highlights that KL design is not merely a training stabilizer but a decisive factor for preserving visual fidelity.
Without KL regularization, generations deteriorate quickly during RL and tend to collapse toward less informative outputs.
More importantly, using a moving reference $p_{\theta_{\mathrm{old}}}$ for the KL penalty (w Step KL), despite being standard in many GRPO/PPO implementations, can further exacerbate degradation in this task, as the reference distribution itself drifts and fails to constrain accumulated mode collapse.
We finally found anchoring the KL term to the fixed SFT policy $p_{\theta_{\mathrm{initial}}}$ provides a stable behavioral prior and consistently maintains trajectory drawing quality.

\subsection{Case Study}

\begin{figure}[h]
  \centering
  \includegraphics[width=\linewidth]{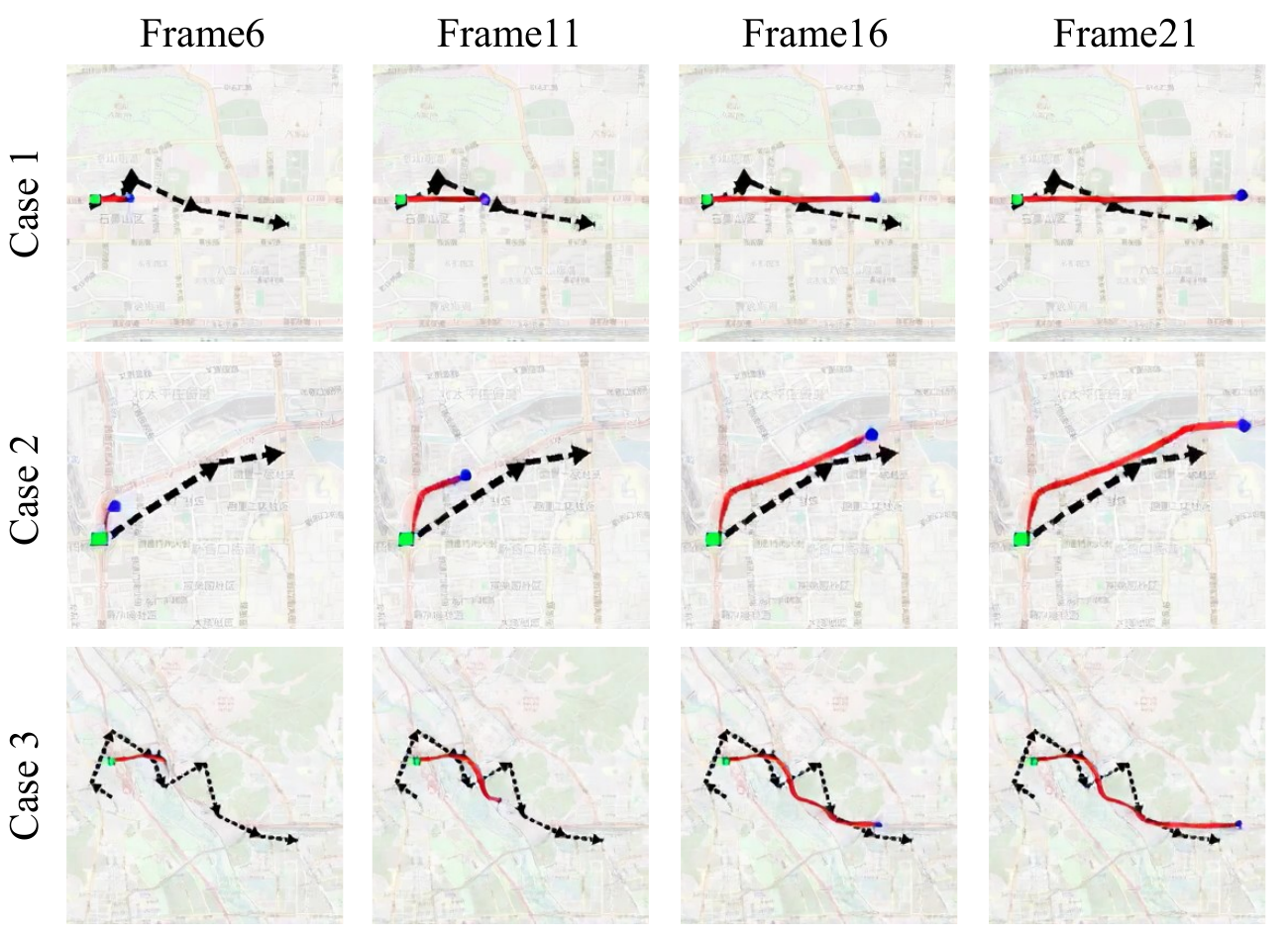}
  \caption{Representative case of Sig2GPS video generation.}
  \label{fig:case}
\end{figure}

In the Figure~\ref{fig:case}, we present three representative scenarios to illustrate our model; Frame~X denotes the $X$-th frame of the generated video.
First, all three examples produce visually realistic vehicle motion along the road network, highlighting the potential of framing Sig2GPS as a video-generation problem.
Second, Case~1 corresponds to driving along a straight road. The model reconstructs this behavior faithfully. Moreover, the trajectory tends to stay on the right-hand side of the roadway, consistent with right-side driving rules.
In Case~2, the route follows a curved road. The model also captures the turning behavior well. Notably, the turning segment progresses more slowly than the straight segment, matching physical intuition.
Case~3 covers a substantially larger and more complex region, yet the model still produces a coherent route that resembles an engineer's map-based reasoning.
Overall, these results suggest that map-visual video generation can effectively denoise and complete coarse signaling traces into road-constrained trajectories across diverse topologies.

\section{Conclusion}

This paper introduces Think Over Trajectory, a map-visual video generation paradigm for Sig2GPS that turns coarse cellular signaling traces into continuous, road-constrained trajectories by drawing in the map domain.
Beyond a strong flow-based SFT initialization on matched signaling-taxi pairs, the proposed Traj-GDPO further aligns generations with fine-grained trajectory criteria through verifiable rewards.
Extensive experiments on real-world data demonstrate consistent gains over both learning-based baselines and a production-grade engineered pipeline, while qualitative results show topology-consistent routes across diverse scenarios.
These findings suggest that keeping both conditioning and prediction in a unified visual map-based space  can offer a practical interface for trajectory data mining, enabling scalable refinement under map constraints.
Future work includes improving robustness under sparse signaling and unseen map styles, as well as extending the framework to broader mobility tasks such as trajectory recovery and long-horizon forecasting.

\newpage 
\bibliographystyle{ACM-Reference-Format}
\bibliography{sample-base}

\begin{figure*}[t]
  \centering
  \includegraphics[width=0.95\textwidth]{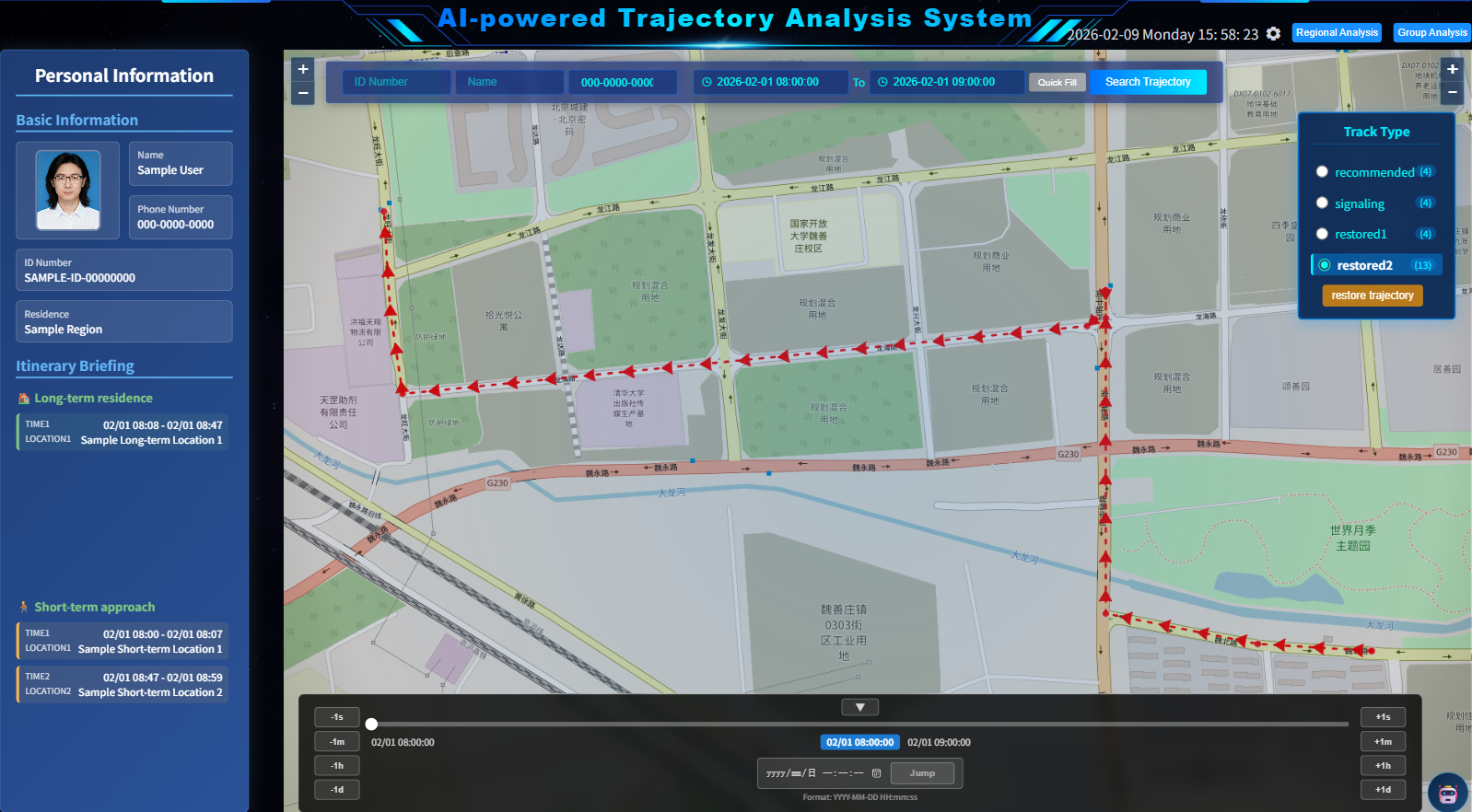}
  \caption{System overview of the deployed Sig2GPS service.}
  \label{fig:system}
\end{figure*}

\newpage 
\appendix

\section{Evaluation Metrics}

Let $\{(p_n,\hat{p}_n)\}_{n=1}^{N}$ denote the set of matched ground-truth and predicted points, and define the point-wise error as $e_n=d_{\mathrm{geo}}(\hat{p}_n, p_n)$.
We report mean absolute error (MAE) and root mean squared error (RMSE):
\begin{equation}
\mathrm{MAE}=\frac{1}{N}\sum_{n=1}^{N}|e_n|,\qquad
\mathrm{RMSE}=\sqrt{\frac{1}{N}\sum_{n=1}^{N}e_n^2}.
\end{equation}

\section{Dataset}

\subsection{Sig2GPS}
\begin{table}[h]
\centering
\caption{Trajectory distance statistics.}
\label{tab:distance_bins_transposed}
\setlength{\tabcolsep}{4.2pt}
\renewcommand{\arraystretch}{1.18}
\begin{tabular}{lcccc}
\toprule
\textbf{Metric} & \textbf{All} & \textbf{Short} & \textbf{Medium} & \textbf{Long} \\
\midrule
Count & 10000 & 3334 & 3333 & 3333 \\
Min (m)  & 157.40 & 157.40 & 2072.35 & 3482.46 \\
Max (m)  & 32331.33 & 2072.24 & 3482.17 & 32331.33 \\
Mean (m) & 3155.13 & 1438.47 & 2711.10 & 5315.60 \\
\bottomrule
\end{tabular}
\end{table}

Here, we provide the statitics of our Sig2GPS Dataset.

\subsection{Next GPS Prediction}
We adopt the Chengdu and Xi’an subsets from the DiDi Chuxing GAIA Open Dataset, which is widely used as a benchmark for Next GPS Prediction. The dataset was released by DiDi through the GAIA initiative for academic research and is irreversibly anonymized to protect user privacy. In the standard trajectory-forecasting setup, the data are organized at the trip/order level: each sample is a time-ordered sequence of GPS observations (e.g., longitude, latitude, and timestamps), reflecting real vehicle movements on an urban road network. Such trajectories provide rich spatiotemporal signals for learning road-topology priors and urban mobility dynamics, making them suitable for next-step location/GPS prediction and trajectory generation evaluation. Our training, validation test set, each has 30000, 1000, 1000 samples to facilatete computation.

\section{Baseline Introduction}

\subsection{Sig2GPS}
- GRU: 
Gated Recurrent Unit (GRU) is a widely used recurrent neural network that models sequential dependencies through update and reset gates. Due to its efficiency and strong performance on generic time-series problems, GRU serves as a standard baseline for capturing temporal patterns in trajectory-like sequences.

- GPT: 
GPT-style models are autoregressive Transformer architectures that leverage self-attention to model long-range dependencies in sequences. As a baseline, GPT provides a strong reference for assessing the effectiveness of large-scale attention-based sequence modeling on our task, especially under complex contextual and long-horizon dependencies.

- MLP:
Multi-Layer Perceptron (MLP) is a feed-forward neural network that predicts outputs from aggregated or flattened input features. Although it does not explicitly model temporal dynamics, it is a simple and stable baseline that helps quantify the benefit brought by dedicated sequential modeling.

- TCN: 
Temporal Convolutional Network (TCN) employs 1D convolutions along the temporal dimension, often with dilated convolutions and residual connections to enlarge the receptive field. TCN is effective at capturing local-to-midrange temporal patterns with high parallelism and is commonly adopted as a competitive baseline for sequence prediction.

- TrajFormer: 
TrajFormer is an efficient Transformer for trajectory sequence modeling that explicitly handles irregular spatio-temporal intervals via a Continuous Point Embedding (CPE) module, and improves scalability with a squeezed self-attention mechanism that compresses the key/value space to reduce attention cost.

- SigFormer: 
SigFormer is a Transformer-based, location-aware model designed for mobile signaling data augmentation via location reconstruction. It leverages self-attention to capture long-range spatio-temporal dependencies and reconstructs missing/incorrect base-station locations from the surrounding signaling context.

\subsection{Next GPS Prediction}
-DeepMove: 
DeepMove is an attentional recurrent network for mobility/next-location prediction from lengthy and sparse trajectories. It combines a multi-modal embedding + recurrent module to model complex transitions, and a historical attention module to capture multi-level mobility periodicity by selectively attending to relevant historical records. 

-GETNext: 
GETNext is a Graph Enhanced Transformer for next-POI prediction. It builds a user-agnostic trajectory flow map to model global POI transitions, learns POI representations with a GNN/GCN, and injects transition signals via a transition attention map, while a Transformer encoder fuses these global patterns with spatio-temporal/context embeddings for prediction. 

-TrajFormer: 
TrajFormer is an efficient Transformer for trajectory modeling that addresses irregular spatio-temporal intervals using Continuous Point Embedding (CPE) and improves scalability with a Squeezed Transformer Encoder that compresses keys/values before attention to reduce computation; it can also use subpath labeling as auxiliary supervision.

-AgentMove:
AgentMove is an LLM-based agentic framework for zero-shot next-location prediction. It decomposes the task and integrates (i) spatial-temporal memory (short-term, long-term, user profile) for individual patterns, (ii) a world knowledge generator for multi-scale urban structure, and (iii) a collective knowledge extractor (graph/tool-based) for shared mobility patterns, followed by a final reasoning step. 

-STAN
STAN (Spatio-Temporal Attention Network) is a bi-layer self-attention model for next-location recommendation. It explicitly models pairwise spatiotemporal effects across all check-ins, enabling interactions between non-adjacent locations and non-consecutive visits, and uses a two-stage attention design to better account for personalized item frequency (PIF). 

-Traj-MLLM:
Traj-MLLM is a training-free MLLM-based framework for region-agnostic, task-adaptive trajectory mining. It segments trajectories into map-anchored sub-trajectories, converts them into interleaved image-text representations with multi-view context, and relies on prompt optimization for flexible task adaptation. 

\end{document}